\documentclass[lettersize,journal]{IEEEtran}
\usepackage{amsmath,amsfonts}
\pdfoutput=1
\usepackage{algorithmic}
\usepackage{algorithm}
\usepackage{array}
\usepackage[caption=false,font=normalsize,labelfont=sf,textfont=sf]{subfig}
\usepackage{textcomp}
\usepackage{stfloats}
\usepackage{url}
\usepackage{verbatim}
\usepackage{graphicx}
\usepackage{cite}
\usepackage{siunitx}
\usepackage{amssymb}
\usepackage{yfonts}
\usepackage{bigstrut,multirow,rotating}
\usepackage{booktabs}

\begin{document}

\title{CBARF: Cascaded Bundle-Adjusting Neural Radiance Fields from Imperfect Camera Poses}

\author{Hongyu Fu, Xin Yu, Lincheng Li, and Li Zhang 
\thanks{Hongyu Fu is with the Department of Electronic Engineering, Tsinghua University, Beijing 100084, China, and also with the NetEase Fuxi AI Lab, Hangzhou 310052, China(e-mail: fhy21@mails.tsinghua.edu.cn).}
\thanks{Xin Yu is with the School of Electronic Engineering and Computer Science, University of Queensland, Brisbane, Australia (e-mail: xin.yu@uq.edu.au).}
\thanks{Lincheng Li is with the NetEase Fuxi AI Lab, Hangzhou 310052, China(e-mail: lilincheng@corp.netease.com).}
\thanks{Li Zhang is with the Department of Electronic Engineering, Tsinghua University, Beijing 100084, China(e-mail: chinazhangli@mail.tsinghua.edu.cn).}
\thanks{Manuscript received April 19, 2021; revised August 16, 2021.}}

\markboth{Journal of \LaTeX\ Class Files,~Vol.~14, No.~8, August~2021}%
{Shell \MakeLowercase{\textit{et al.}}: A Sample Article Using IEEEtran.cls for IEEE Journals}

\IEEEpubid{0000--0000/00\$00.00~\copyright~2021 IEEE}

\maketitle




\begin{abstract}

Existing volumetric neural rendering techniques, such as Neural Radiance Fields (NeRF), face limitations in synthesizing high-quality novel views when the camera poses of input images are imperfect. 
To address this issue, we propose a novel 3D reconstruction framework that enables simultaneous optimization of camera poses, dubbed CBARF (Cascaded Bundle-Adjusting NeRF).
In a nutshell, our framework optimizes camera poses in a coarse-to-fine manner and then reconstructs scenes based on the rectified poses.
It is observed that the initialization of camera poses has a significant impact on the performance of bundle-adjustment (BA). Therefore, we cascade multiple BA modules at different scales to progressively improve the camera poses.
Meanwhile, we develop a neighbor-replacement strategy to further optimize the results of BA in each stage. 
In this step, we introduce a novel criterion to effectively identify poorly estimated camera poses. Then we replace them with the poses of neighboring cameras, thus further eliminating the impact of inaccurate camera poses.
Once camera poses have been optimized, we employ a density voxel grid to generate high-quality 3D reconstructed scenes and images in novel views. 
Experimental results demonstrate that our CBARF model achieves state-of-the-art performance in both pose optimization and novel view synthesis, especially in the existence of large camera pose noise.

\end{abstract}



\section{Introduction}
\IEEEPARstart{T}{hree} Dimensional Reconstruction \cite{geiger2011stereoscan,mouragnon2006real,kang2020review,izadi2011kinectfusion,pollefeys2008detailed,schonberger2016structure} and Novel View Synthesis \cite{avidan1997novel,park2017transformation,riegler2020free,zhou2018stereo,mildenhall2021nerf} are essential tasks in computer vision \cite{choy20163d}. They aim to reconstruct 3D scenes from the given 2D RGB images and render photo-realistic images in novel views.
Inspired by the success of Neural Radiance Fields (NeRF) \cite{mildenhall2021nerf}, volumetric neural rendering methods have gained significant popularity in the field of 3D reconstruction in recent years.
However, one limitation of existing methods is the requirement for accurate camera poses corresponding to each input image. In other words, when the input camera poses contain noise or are even completely unknown, these methods might fail to reconstruct scenes or generate high-quality novel views.

The recent work BARF \cite{lin2021barf} attempts to solve camera registration and scene reconstruction jointly. BARF can be considered as a variant of photometric Bundle-Adjustment(BA) \cite{triggs1999bundle,wu2011multicore,agarwal2010bundle,zach2014robust,engels2006bundle} with view synthesis serving as a proxy objective.
BARF can effectively correct camera poses with moderate noise and reconstruct scenes when camera poses lie in restricted 3D space, \emph{e.g.}, sharing similar orientations and lying on a common 2D plane.
However, when registering camera poses in a 3D free space, BARF might fail due to the increased optimization difficulty of the joint estimate of camera poses and scene reconstruction.
We observe that even when cameras are distributed in a 3D hemispherical space and face an object positioned at the center, BARF cannot handle camera pose noise and produces inferior reconstruction results.
Moreover, in some cases, input images do not have the corresponding camera pose information. For instance, when COLMAP~\cite{schonberger2016structure} might fail to estimate camera poses for some images, these images will not be used for scene reconstruction. This would lead to inferior reconstruction results, such as some parts of scenes are missing.

\IEEEpubidadjcol

\begin{figure}[t]
	\centering
	\includegraphics[width=1.0\linewidth, trim=3mm 3mm 3mm 3mm, clip]{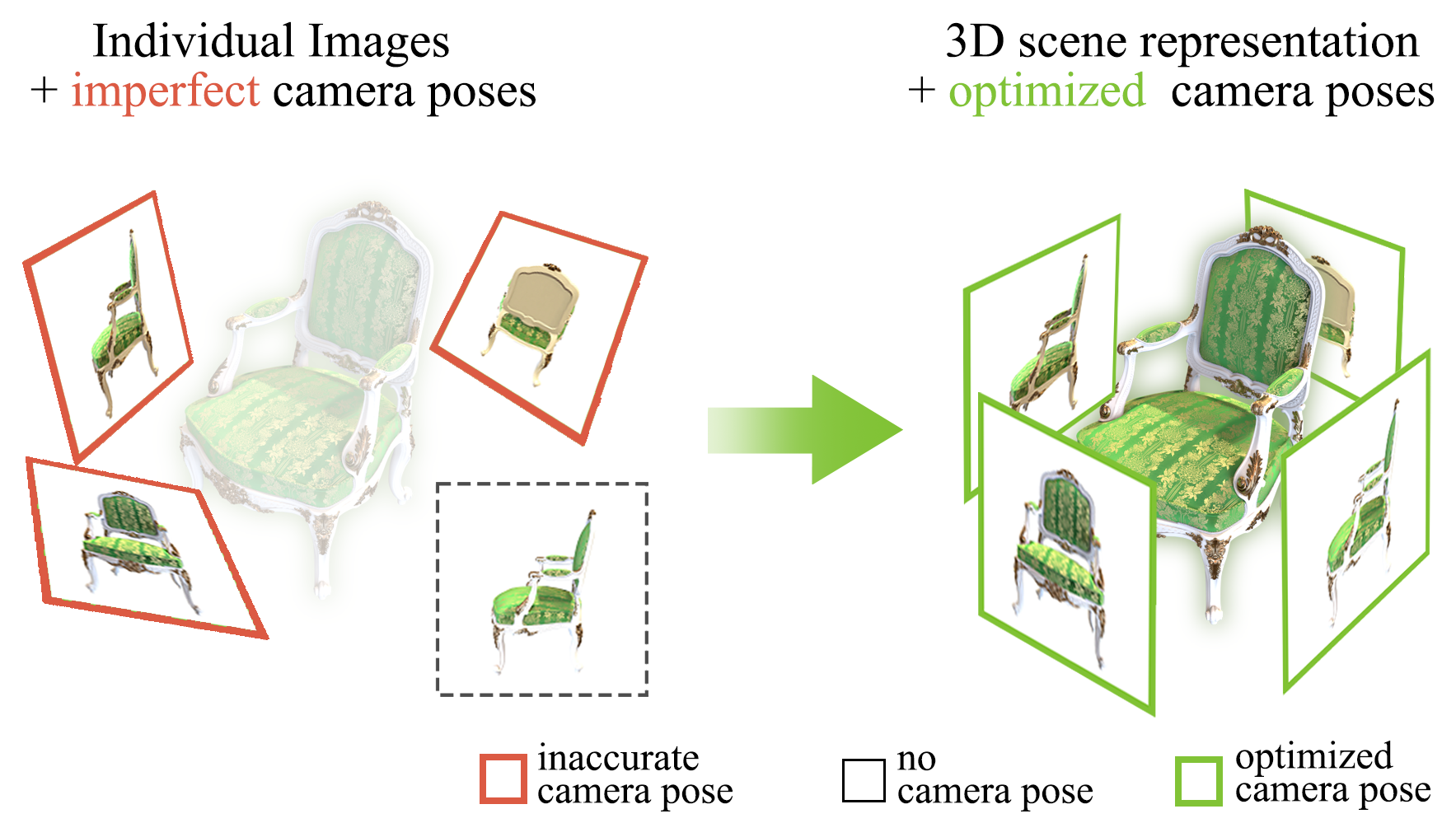}
	\vspace{-1em}
	\caption{Learning 3D scene representations relies on accurate camera poses of input images. However, coping with inaccurate or incomplete camera poses imposes a challenge. Our proposed CBARF tackles this problem by effectively reducing large camera pose noise and estimating missing camera poses.
} 
\label{fig.task}
\end{figure}

To address these issues, we propose Cascaded Bundle-Adjusting Neural Radiance Fields (CBARF), a novel approach to reconstructing scenes from inaccurate or partially unknown camera poses.
Our CBARF model adopts a coarse-to-fine manner. In each scale, CABRF first updates camera poses by a BA module. Subsequently, we design a novel criterion to identify poorly estimated poses that still have not been rectified after BA optimization. We then introduce a neighbor-replacement strategy to update these inaccurate poses.

Specifically, we find that optimizing camera poses with excessive iterations is computationally costly and does not lead to better performance. Thus, we introduce a compact BA module by modifying the basic BA module to accelerate the pose optimization process.
Due to the substantial influence of the initialization state of camera poses on BA, we adopt the preceding pose estimation results as the initialization for the subsequent BA module.
As a result, we cascade multiple compact BA modules in series, forming the backbone of our CBARF model.
The number of cascades in our model is adaptive to avoid insufficient or excessive optimization rounds.

To further enhance the performance of the cascaded BA, we introduce a neighbor-replacement strategy between each pair of BA modules. This strategy involves replacing inaccurate camera poses with poses from neighboring viewpoints.
Due to the absence of ground-truth camera poses, we design a novel criterion to identify potentially inaccurate camera poses based on the quality of rendered images in the corresponding views.
In addition, we incorporate non-maxima suppression \cite{neubeck2006efficient,hosang2017learning,hosang2016convnet} to enhance the identification of inaccurate poses. 
The final refined poses are provided into a density voxel grid \cite{sun2022direct}, facilitating the generation of high-quality rendered images for the purpose of result comparison.
We conducted a comprehensive evaluation and comparison of our approach on the NeRF-synthetic \cite{mildenhall2021nerf} and BlendedMVS \cite{yao2020blendedmvs} datasets. Our results demonstrate that our approach achieves a new state-of-the-art performance in optimizing camera poses from noisy or insufficient initial estimates.

Overall, the contributions of our work are summarized as follows:
\begin{itemize}
\item We propose a robust coarse-to-fine 3D reconstruction framework that effectively optimizes camera poses in the presence of significant noise. Our model exhibits the capability of handling images with noisy camera pose information.

\item We demonstrate that the initialization of camera poses is crucial for bundle-adjustment (BA) performance, and we propose the cascaded BA to progressively refine the inaccurate camera poses.

\item We propose a neighbor-replacement strategy to improve the optimization process by identifying and replacing inaccurate camera poses with the poses of their neighboring cameras.

\end{itemize}


\section{Related Work}

{\noindent {\bf Structure from Motion: }}
Structure from motion (SfM) system \cite{cernea2020openmvs,schonberger2016structure,hartley2004multiple,ozyecsil2017survey,ullman1979interpretation,andersen1998perception,bao2011semantic}, such as the COLMAP \cite{schonberger2016structure} aims to estimate the camera poses and recover the 3D structure from the given a set of input images.
Since SfM only needs RGB images without any pose or depth information as input, it has been widely used to reconstruct sparse point clouds and recover camera poses. 
Many works on SfM achieve great success such as COLMAP \cite{schonberger2016structure} and OpenMVS \cite{cernea2020openmvs}.
However, most SfM systems rely on detecting and matching distinctive key-points. Thus, they may fail to reconstruct scenes, especially in regions with less texture or repetitive patterns.

\vspace{1em}
{\noindent {\bf Neural Radiance Field: }}
With the continuous advancement of deep learning techniques, neural networks have found wide applications in 3D reconstruction and novel view synthesis \cite{9174748,8962030,9417684,6359953,10113194,9113759,9339999,9759982,9891833,mescheder2019occupancy,han2019image,zhang20213d}. NeRF (Neural Radiance Field \cite{mildenhall2021nerf}) is one such technique. NeRF aims to learn scene representation inside an MLP and synthesize novel views directly via differentiable volume rendering. Due to its photo-realistic rendering capabilities, NeRF has gained a lot of attention in various fields, such as high-quality head reconstruction \cite{9537697,10229247}. Many researchers have explored ways to improve its performance and address its weaknesses \cite{yu2021pixelnerf,fridovich2022plenoxels,martin2021nerf,sun2022direct,barron2021mip,pumarola2021d,lin2021barf,chng2022gaussian,Truong_2023_CVPR,wang2021nerf}.
Some works aim to predict a continuous neural scene representation from a sparse set of input views. 
PixelNeRF \cite{yu2021pixelnerf} employs a fully convolutional approach for processing image inputs, enabling the network to be trained across multiple scenes and learn a scene prior. This facilitates generating novel view synthesis conditioned on a limited number of input images.
Many researchers have attempted to accelerate the training process of NeRF.
Plenoxels \cite{fridovich2022plenoxels} achieves comparable performance to NeRF but with a significant speed improvement, being approximately 100 times faster.
It utilizes a sparse voxel grid representation, where each voxel is associated with density and spherical harmonic coefficients.
DVGO \cite{sun2022direct} further advances NeRF and 3D scene representation. By using a voxel grid representation rather than an MLP, DVGO can significantly accelerate the rendering process compared to traditional methods.
However, those existing volumetric neural rendering methods require a set of images with accurate poses as input. They may fail to synthesize high-quality novel views when camera poses contain some noise. 

\vspace{1em}
{\noindent {\bf Extended NeRF with Inaccurate Poses: }}
Several works aim to reduce the reliance on highly accurate camera poses.
BARF \cite{lin2021barf} jointly optimizes registration and reconstruction from inaccurate camera poses (inward-facing datasets), or even unknown poses (forward-facing datasets).
GARF \cite{chng2022gaussian} presents a new positional embedding-free neural radiance field architecture with Gaussian activation to solve the joint problem of reconstruction and pose estimation. SPARF \cite{Truong_2023_CVPR} exploits multi-view geometry constraints to jointly learn the scene representation and refine the camera poses from sparse viewpoints.
NeRF-- \cite{wang2021nerf} optimizes camera poses as learnable parameters with NeRF training through a photometric reconstruction on forward-facing datasets.

However, these methods still fail to learn scene representations when some camera poses contain severe noise.
Therefore, our proposed CBARF is proposed to reconstruct 3D scenes when images only have inaccurate or insufficient camera poses.


\begin{figure*}[t]
	\centering
	\includegraphics[width=1.0\linewidth]{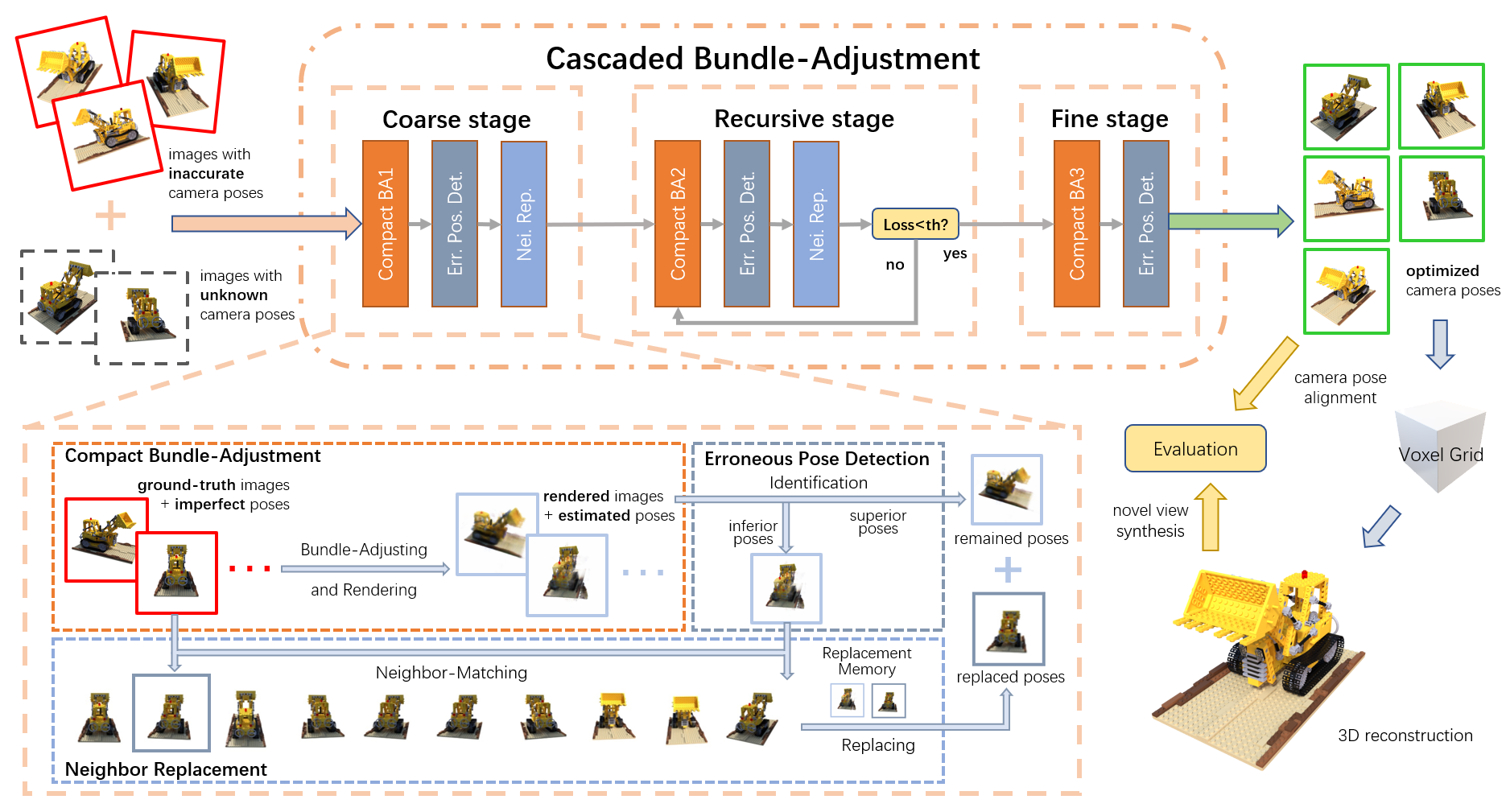}
	\vspace{-1em}
	\caption{The overview of our proposed CBARF. At each stage of the cascaded BA, we use a compact BA module to rectify camera poses and generate rendered images. Subsequently, we identify the inferior rendered images to select potentially inaccurate camera poses and update these erroneous poses by their nearest poses found across all views. The updated poses are then fed into the next stage for further optimization. During the recursive stage, we employ a loop detection technique to automatically determine the number of compact BA modules. In the final stage, the optimized poses with reduced noise are checked again to remove any incorrect poses. The remaining poses are then input into a density voxel grid for high-quality reconstruction.
} 
\label{overview}
\end{figure*}

\section{Proposed Method}

This work addresses the challenge of synthesizing novel views under conditions where certain input camera poses contain significant noise or are even unknown. 
Thus, we propose CBARF, a novel framework incorporating several simple yet effective strategies, to optimize camera poses and learn 3D scene representations. 
In this section, we first introduce the cascaded BA in Sec. \ref{cba}. Similar to other gradient descent algorithms, cascaded BA is prone to over-fitting when the initial camera poses have significant errors. 
We then present the neighbor-replacement strategy in two separate parts, detailing the process of identification and replacement of erroneous camera poses.
In Sec. \ref{sec3.2}, we design a novel criterion to detect erroneous camera poses arising from over-fitting during the BA process.
Since ground-truth poses are unknown, this method identifies the potentially inaccurate poses based on the quality of their corresponding rendered images.
We introduce the procedure for replacing these identified erroneous poses in Sec. \ref{sec3.3}. This technique rectifies inaccurate camera poses by replacing them with poses of neighboring cameras.

\subsection{Cascaded BA}

\label{cba}

Employing bundle-adjustment methods \cite{triggs1999bundle,wu2011multicore,agarwal2010bundle,zach2014robust,engels2006bundle,lin2021barf} under conditions where the initial camera poses deviate significantly from ground-truth may result in broken 3D structure and failure pose estimation. The sub-optimal performance of some BA models such as BARF \cite{lin2021barf} may be attributed to the over-fitting of neural radiance field network. 
Specifically, BARF generates novel view images by an MLP and uses the synthesized images as the proxy objective to update camera poses in an alternating manner. 
However, when some input poses contain large deviations, it may learn an incorrect scene representation, leading to an erroneous optimization and preventing the model from self-correcting. 
In essence, the BA model heavily relies on accurate initial poses to establish a reliable starting point for reconstruction and optimization.

In some other optimization tasks \cite{dollar2010cascaded,jarrett2009best,zamir2021multi}, multi-stage structures are employed for iterative refinement and improved performance. 
Inspired by this concept, we propose cascaded BA, a multi-stage framework incorporating several compact BA modules connected in series.
In cascaded BA, the camera poses estimated from the previous stage are utilized as the initialization for the subsequent stage. In this way, the model coarsely eliminates cumulative errors and results in more accurate pose estimation for the subsequent process. 
The compact BA module is based on BARF \cite{lin2021barf}. However, BARF takes much training time to complete the reconstruction. Since high-quality reconstruction is not essential for coarsely optimizing camera poses, we reduce the training iterations of the compact BA module. We also adjust the learning rate to match the different training stages. Consequently, We adopt a coarse-to-fine manner consisting of coarse, recursive, and fine stages (Fig. \ref{overview}).
The number of compact BA modules is adaptive and depends on the characteristics of the datasets. Specifically, we employ a loop detection technique in the recursive stage to assess the current optimization effectiveness and determine the number of compact BA modules. This helps us avoid insufficient or excessive optimization rounds, ensuring an optimal balance for the performance of the model.

\begin{figure}[t]
	\centering
	\includegraphics[width=1.0\linewidth, trim=3mm 3mm 3mm 3mm, clip]{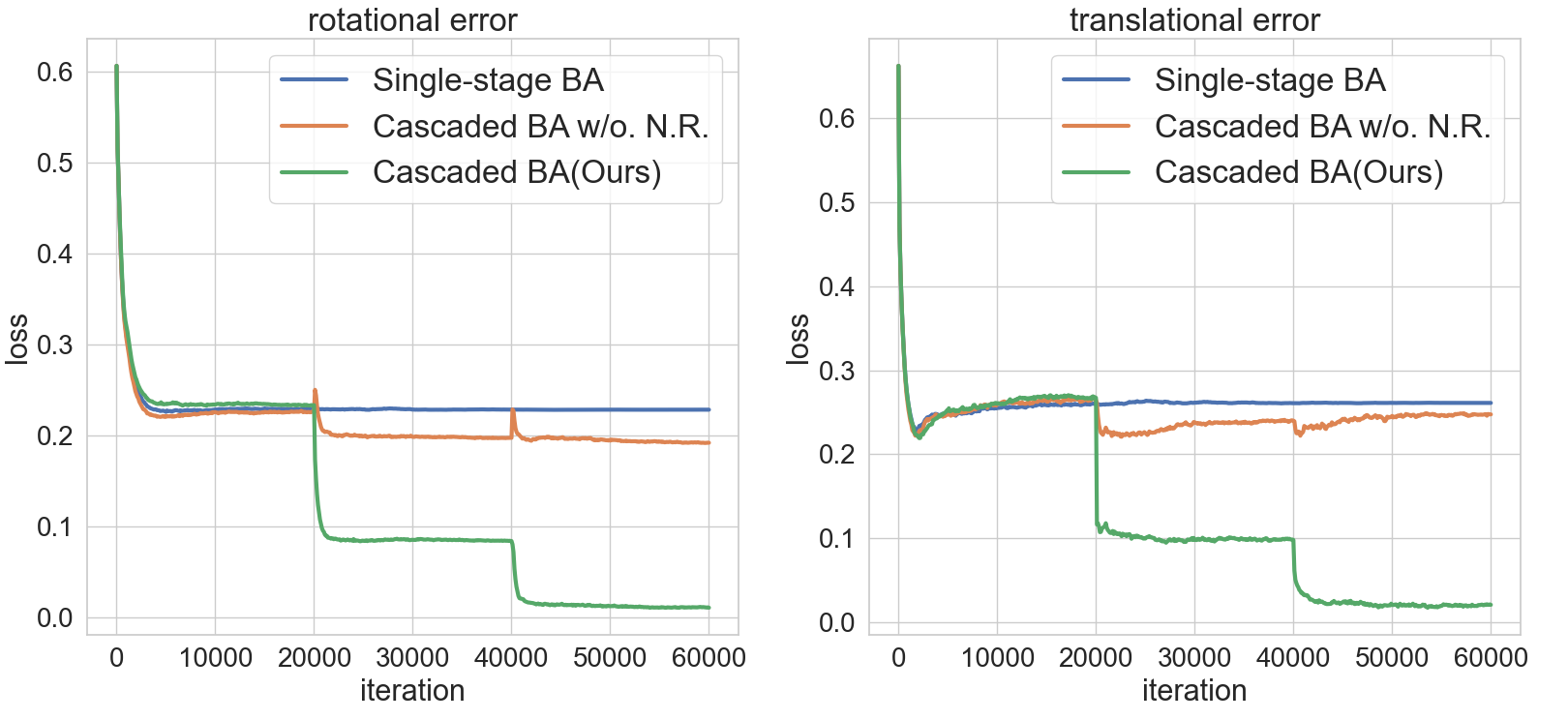}
	\vspace{-1em}
	\caption{Comparison between the cascaded BA and the single-stage BA. In this figure, the cascaded BA consists of three compact BA modules, with each module set to 20k iterations. The single-stage BA is configured with 60,000 iterations, matching the total number of iterations used in the cascaded BA. The red line represents the cascaded BA with neighbor-replacement at each cascaded node, while the green line represents the cascaded BA without neighbor-replacement. The blue line represents the single-stage BA. The single-stage BA quickly falls into a sub-optimal solution, while the cascaded BA exhibits fluctuations at the cascaded nodes, resulting in a better final result. In particular, the red curve exhibits a steeper descent at each cascaded node compared to the green curve. This observation suggests that the neighbor-replacement technique notably enhances the model's performance. 
} 
\label{fig.pose_err}
\end{figure}

Our experiments show that multi-stage BA rectifies inaccurate poses more efficiently than single-stage BA during the same training time. 
As shown in Fig. \ref{fig.pose_err}, the cascaded BA without neighbor-replacement (indicated by the green curve) reduces more amount of camera pose noise than single-stage BA (indicated by the blue curve). The single-stage BA quickly falls into a sub-optimal solution, while the cascaded BA exhibits an improved optimization result. Moreover, the red curve, cascaded BA with neighbor-replacement (Sec. \ref{sec3.3}), demonstrates a more significant reduction in noise.

\subsection{Erroneous Pose Detection}
\label{sec3.2}

\begin{figure}[t]
	\centering
	\includegraphics[width=1.0\linewidth]{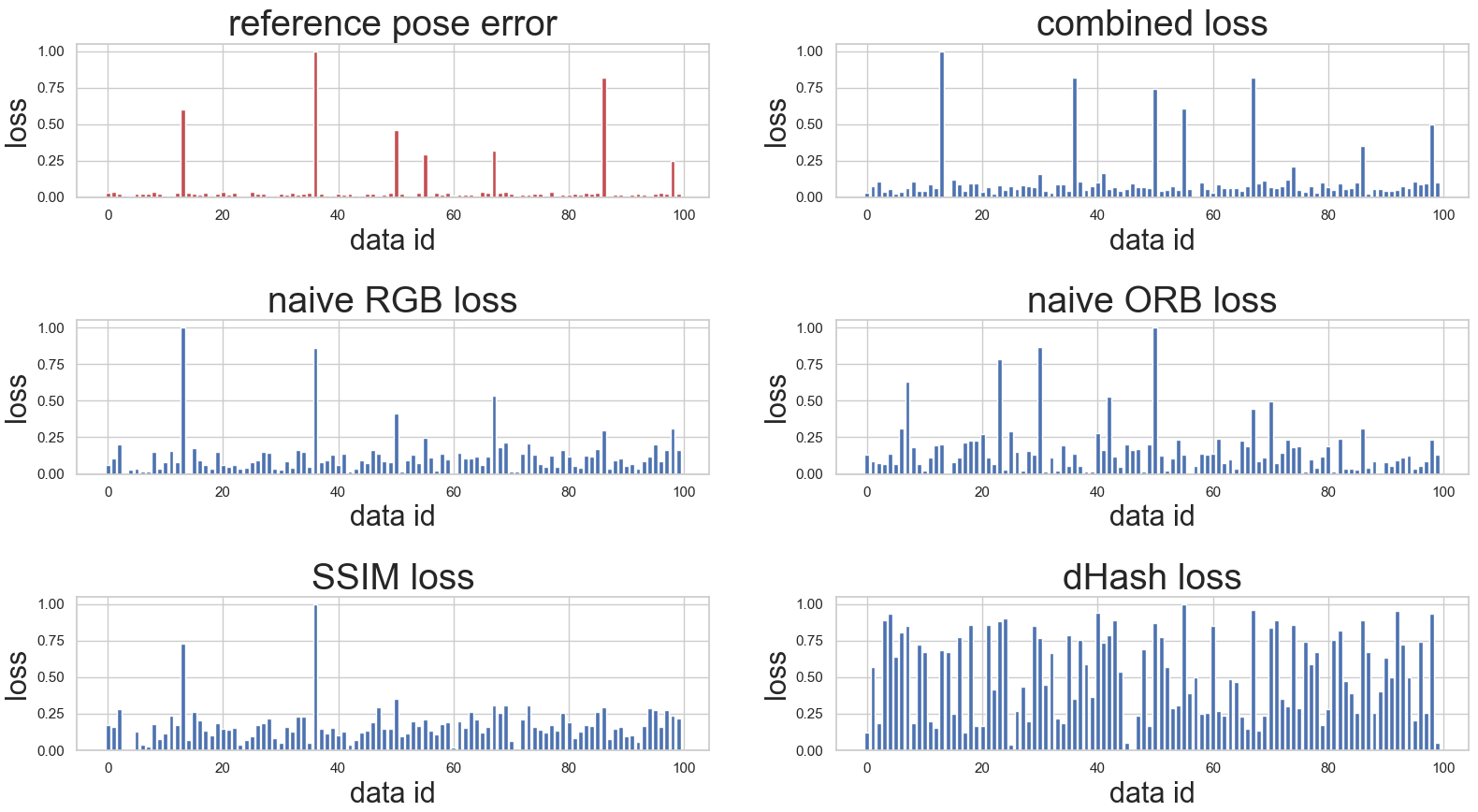}
	\vspace{-1em}
	\caption{Comparison between several evaluation methods. The x-axis of each chart represents the identification numbers of different views, while the y-axis represents normalized error values. The first chart (in red) illustrates the distribution of camera pose errors at different views, while the remaining charts display the distribution of rendered image errors calculated by various image evaluation methods. Compared to other traditional loss, our combined loss exhibits a stronger correspondence between rendered image errors and camera pose errors. This demonstrates the superior capability of our method to identify inaccurate camera poses.}
\label{fig.eva}
\end{figure}

This module aims to overcome the challenge of identifying inaccurate camera poses without ground-truth. It involves generating view synthesis using the current pose estimates and accessing the quality of the rendered images. Inferior rendering quality consistently indicates inaccuracies in the corresponding camera poses.
However, we observe that rendering errors caused by inaccurate camera poses are prone to be confused with noise introduced by the model, especially when using conventional image evaluation algorithms \cite{huynh2008scope,wang2004image,zhang2018unreasonable} for evaluation.
Additionally, even minor inaccuracies in camera poses can result in pixel-level displacements in the rendered images. As a result, conventional image evaluation methods predominantly based on pixel comparisons might consequently lead to erroneous assessments.
Consequently, we introduce a novel criterion primarily based on ORB key-point \cite{rublee2011orb} to overcome the disadvantages of conventional evaluation methods. Additionally, we design several supplementary strategies to improve the accuracy and reliability of the criterion. As illustrated in Fig. \ref{fig.eva}, our combined criterion exhibits superior correlation with camera pose errors when compared to other conventional image evaluation methods.

ORB (Oriented FAST and Rotated BRIEF) \cite{rublee2011orb} is a feature detection and description algorithm used in computer vision and image processing. It is similar to the SIFT (Scale-Invariant Feature Transform) \cite{lowe2004distinctive} algorithm, but is designed to be faster and more efficient. We find that ORB is suitable for finding and matching key-points between rendered images and reference images in our task. 
Additionally, we propose several supplementary techniques to enhance the accuracy of key-point matching.

\begin{itemize}
\item {\bf{K-Nearest Neighbors:}} In order to effectively identify the matched key-points between reference images and rendered images, we employ the K-Nearest Neighbors (KNN) algorithm \cite{cover1967nearest}. 
For each key-point in the rendered image, we compute its feature distance with all the key-points in the reference image. Subsequently, we identify the point with the shortest distance as a potential match and compare its feature distance with others. A dependable match generally shows a significantly shorter feature distance compared to non-matching points. Consequently, potential matches without a notably shorter feature distance are disregarded.

\item {\bf{Bidirectional Check:}} We use a bidirectional check \cite{hao2013fast,wu2022image} between rendered images and reference images to further improve the accuracy of the key-point matching. 
In detail, for each key-point $p_i$ in the rendered image, we conduct a search to find the best matching point $p_m$ in the reference image. Additionally, we traverse the rendered image to confirm whether $p_i$ is also the best matching point for $p_m$. We only retain the reliable matches, where two points serve as each other's best matching points.

\item {\bf{Coordinate Constraints:}} In some cases, inaccurate camera poses result in scenes being rendered from an incorrect viewpoint, while still generating visually high-quality images. In the overlapping regions of the scene captured from different viewpoints, there are numerous shared key-points at different pixel positions. Since the matching method relies on feature distances rather than pixel positions, these displaced key-points may be incorrectly matched. This results in overlooking some inferior rendered images caused by perspective errors.
To address this issue, we incorporate coordinate constraints for key-point pairs. Specifically, when a pair of matched key-points exhibits a significant coordinate separation, we classify them as unreliable matches and discard them to exclude perspective errors.

\end{itemize}

To address the issue of insufficient ORB key-points, we supplement the criterion with RGB-MSE. MSE (Mean Squared Error) is a common loss function used in regression problems \cite{wang2009mean}. It measures the average squared difference between the predicted output and ground-truth.
However, directly using RGB-MSE in our task yields unsatisfactory results. We observe that variations in the foreground-to-background ratio affect the comparability of MSE values across different viewpoints. Therefore, we introduce a compensating factor to address this issue. The revised MSE can be described as
\begin{equation}
MSE_c = { \sqrt{\frac{N}{N_f}} \cdot \frac{1}{N} \sum_{i=1}^{N} (y_i - \hat{y}_i)^2},
\end{equation}
where $N_f$ represents the number of pixels in the foreground of the reference image, and $N$ represents the total number of pixels. $y_i$ represents the pixel value of ground-truth image while $\hat{y}_i$ represents the rendered image.

In the final step, we utilize the combined criterion to assign scores for all the rendered images obtained from the current pose estimates. By analyzing these scores, we identify the low-quality images and their corresponding camera poses. The poses are considered to contain a high level of noise and will be further optimized in neighbor-replacement \ref{sec3.3}.

\subsection{Neighbor-Replacement}
\label{sec3.3}
The neighbor-replacement technique involves replacing the camera poses identified as low-quality in sec. \ref{sec3.2} with their respective neighbors. 
Specifically, we first denote $\mathbb{Q} = \{ q_k |k = 1,2,...,N \}$ as all views from the input set.
Note that we have labeled each $q_k$ as either 'superior' or 'inferior' in sec. \ref{sec3.2}. We denote them as $q_s \in \mathbb{Q}_{s}$ and $q_i \in \mathbb{Q}_{i}$, respectively.

However, we observe that inaccurate camera poses have a detrimental effect on the rendering not only in their corresponding viewpoints but also in neighboring viewpoints. This interference can arise from the model learning incorrect scene representations in the relevant regions. Consequently, our proposed criterion in sec. \ref{sec3.2} may erroneously identify some accurate camera poses as low-quality due to the broken scene representations caused by neighboring inaccurate poses. To address this issue, we introduce non-maxima suppression \cite{neubeck2006efficient,hosang2017learning,hosang2016convnet} after identification.  
Specifically, when multiple neighboring camera poses are assigned low scores, it is probable that only the camera pose with the lowest score is inaccurate. Other camera poses are mistakenly labeled as low-scoring poses.
Thus, we update the classification of 'superior' and 'inferior' in set $\mathbb{Q}$ by discarding misidentified camera poses. We denote the updated classifications as $\mathbb{\widetilde{Q}}_{s}$ and $\mathbb{\widetilde{Q}}_{i}$, respectively. 

For each inferior view $\widetilde{q}_i$ in set $\mathbb{\widetilde{Q}}_{i}$, we search for the nearest neighbor view $\widetilde{q}_{simi}$ in the set $\mathbb{Q}$. We adopt the combined criterion introduced in Sec. \ref{sec3.2} to measure and rank the similarity between $\widetilde{q}_i$ and each $q_i$. We additionally perform matching on the rotated images with their corresponding camera poses equivalently rotating \cite{shorten2019survey}. This approach expands the search space and improves the accuracy of neighbor-matching results.
The camera poses in set $\mathbb{\widetilde{Q}}_{i}$ are then replaced with their nearest neighbor in set $\mathbb{\widetilde{Q}}_{s}$. 
As a result, we obtain an updated set of camera pose estimates denoted as $\mathbb{\widetilde{Q}} = \{ \widetilde{q}_k |k = 1,2,...,N \}$. The $\mathbb{\widetilde{Q}}$ is more reliable for further optimization.

However, the effectiveness of replacement depends on the accuracy of neighbor-matching. A mismatched neighboring camera pose may be replaced rapidly, leading to eventual optimization failure. Even with the enhancement measures in Sec \ref{sec3.2}, some matching results may still be incorrect. Hence, we introduce a replacement-memory technique to prevent redundant erroneous replacements. For each erroneous viewpoint, we keep track of the pose used for replacement and skip poses that have already been utilized. This prevents the optimization process from getting stuck due to inaccurately matched neighboring camera poses.

\begin{figure}[t]
	\centering
	\includegraphics[width=1.0\linewidth]{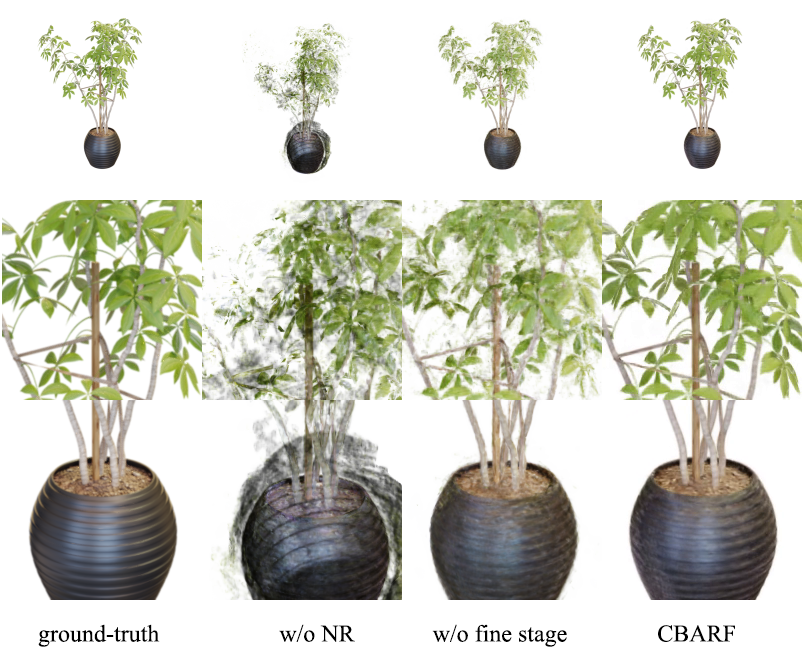}
	\vspace{-1em}
	\caption{Qualitative results of CBARF with different module compositions. We test CBARF without the neighbor-replacement or the fine stage, and visualize the image synthesis at estimated poses. CBARF achieves comparable synthetic quality to the ground-truth, indicating successful pose optimization. On the other hand, the absence of the neighbor-replacement or the fine stage resulted in suboptimal registration, leading to synthesis artifacts in rendered images.
}
\label{fig.ab}
\end{figure}

By replacing the inferior camera poses with more accurate ones, the neighbor-replacement technique improves the initialization effect of each phase in the cascaded BA and significantly enhances the overall performance of the model. As shown in Fig. \ref{fig.pose_err}, the cascaded BA with neighbor-replacement (indicated by the red curve) effectively reduces the camera pose noise after each cascaded node, ultimately reaching an extremely low level. Improved camera pose estimation results in higher rendering quality. Fig. \ref{fig.ab} illustrates the rendering performance of CBARF with different module compositions. Benefiting from the Neighbor-Replacing module and the coarse-to-fine manner, CBARF achieves comparable rendering performance to the ground-truth.


\section{Experiments}

In this section, we first validate the effectiveness of our proposed CBARF in pose registration and view synthesis when dealing with noisy camera poses. Subsequently, we evaluate CBARF's capability of learning scene representations from incomplete camera pose data.

\subsection{Optimizing Noisy Camera Poses}
\label{exp1}

\vspace{2mm}
\noindent \textbf{Experimental settings.}
We conduct evaluations on the NeRF-synthetic dataset \cite{mildenhall2021nerf} with image resolutions of 800$\times$800. 
The camera poses are described by the Lie group $\mathnormal{SE}(3)$ \cite{mishra2013lie}. The Lie group can be defined as
\begin{equation}
SE(3) = \left\{\begin{bmatrix} \mathbf{R} & \mathbf{t} \\
\mathbf{0}^T & 1 \end{bmatrix} \in \mathbb{R}^{4 \times 4}|\mathbf{R} \in SO(3),\mathbf{t} \in \mathbb{R}^{3}\right\},
\end{equation}
where the $\mathbf{R}$ represents the rotation matrix and $\mathbf{t}$ represents the translation matrix. 
$\mathnormal{SO}(3)$ \cite{lee2018introduction} refers to the special orthogonal group in three dimensions and often be used to describe the rotations.
Note that we drop the last row in the Lie group, so the camera poses $\mathbf{P}$ is
\begin{equation}
\mathbf{P} = \left\{ \begin{bmatrix} \mathbf{R}_p & \mathbf{t}_p \end{bmatrix}\in \mathbb{R}^{3 \times 4} |\mathbf{R}_p \in SO(3),\mathbf{t}_p \in \mathbb{R}^{3}\right\}.
\end{equation}

To simulate inaccurate camera poses, we introduce noise $\mathfrak{n}\in\mathfrak{se}(3)$ by generating 6-dimensional random normal distribution noise based on the Lie algebra \cite{mishra2013lie}.
We set the noise coefficient to 0.35 for our method, while it is set to 0.15 in BARF \cite{lin2021barf}. 
The noise $\mathfrak{n}$ in our method can be described as
\begin{equation}
\label{035noise}
\mathfrak{n}= 0.35\mathcal{N}(\mathbf{0}, \mathbf{I})\in \mathbb{R}^6.
\end{equation}
It corresponds to an average deviation of 30.4° in rotation and 0.56 in translation. 
We then transform the noise $\mathfrak{n}$ into the camera transform matrix $\mathbf{N} = \begin{bmatrix} \mathbf{R}_n & \mathbf{t}_n \end{bmatrix}$.
We compose it with the reference camera pose $\mathbf{P}$ to get the imperfect camera poses $\mathbf{\widetilde{P}}$ as
\begin{equation}
\mathbf{\widetilde{P}} =  \begin{bmatrix} \mathbf{R}_p\mathbf{R}_n & \mathbf{t}_p + \mathbf{t}_n \end{bmatrix}.
\end{equation}

In particular, we optimize the camera poses by training the camera refine parameters $\mathfrak{p}\in\mathfrak{se}(3)$. We then convert it into camera transform matrix $\mathbf{P}_r = \begin{bmatrix} \mathbf{R}_r & \mathbf{t}_r \end{bmatrix}$ to compose with the $\mathbf{\widetilde{P}}$. In this way, we obtain the refined camera poses as
\begin{equation}
\mathbf{Q} =  \begin{bmatrix} \mathbf{R}_p\mathbf{R}_n\mathbf{R}_r & \mathbf{t}_p + \mathbf{t}_n + \mathbf{t}_r \end{bmatrix}.
\end{equation}

\vspace{2mm}
\noindent \textbf{Implementation Details.}
In this paper, we propose a coarse-to-fine structure (Fig. \ref{overview}) to optimize camera poses and reconstruct 3D scenes. 
For each compact BA module, we choose the BARF network \cite{lin2021barf} as our backbone. We follow the architectural settings from the original BARF with some modifications and adopt a coarse-to-fine strategy for each optimization stage. Specifically, in the coarse stage and the recursive stage, the iteration for the compact BA module is set to 20k, while in the fine stage it is 200k. To further improve training efficiency, we reduce the image sizes by half during the pose optimization stage, resulting in resolutions of 400$\times$400.
Similar to BARF\cite{lin2021barf} and NeRF \cite{mildenhall2021nerf}, we employ exponential interpolation \cite{smith2017cyclical,carlson1995fast} to calculate a gradually decreasing learning rate. Additionally, we introduce a modulation factor to determine the degree of deviation from the initial value. A larger modulation factor biases the overall learning rate more towards the initial high value. Setting the modulation factor to 1.0 corresponds to using the original exponential learning rate. In the three stages of the cascaded BA, the modulation factors were set to 10.0, 3.0, and 1.0, respectively.
In the Voxel Grid Module, DVGO \cite{sun2022direct} is employed to generate synthetic images with optimized camera poses for evaluation.

\begin{figure}[t]
	\centering
	\includegraphics[width=1.0\linewidth]{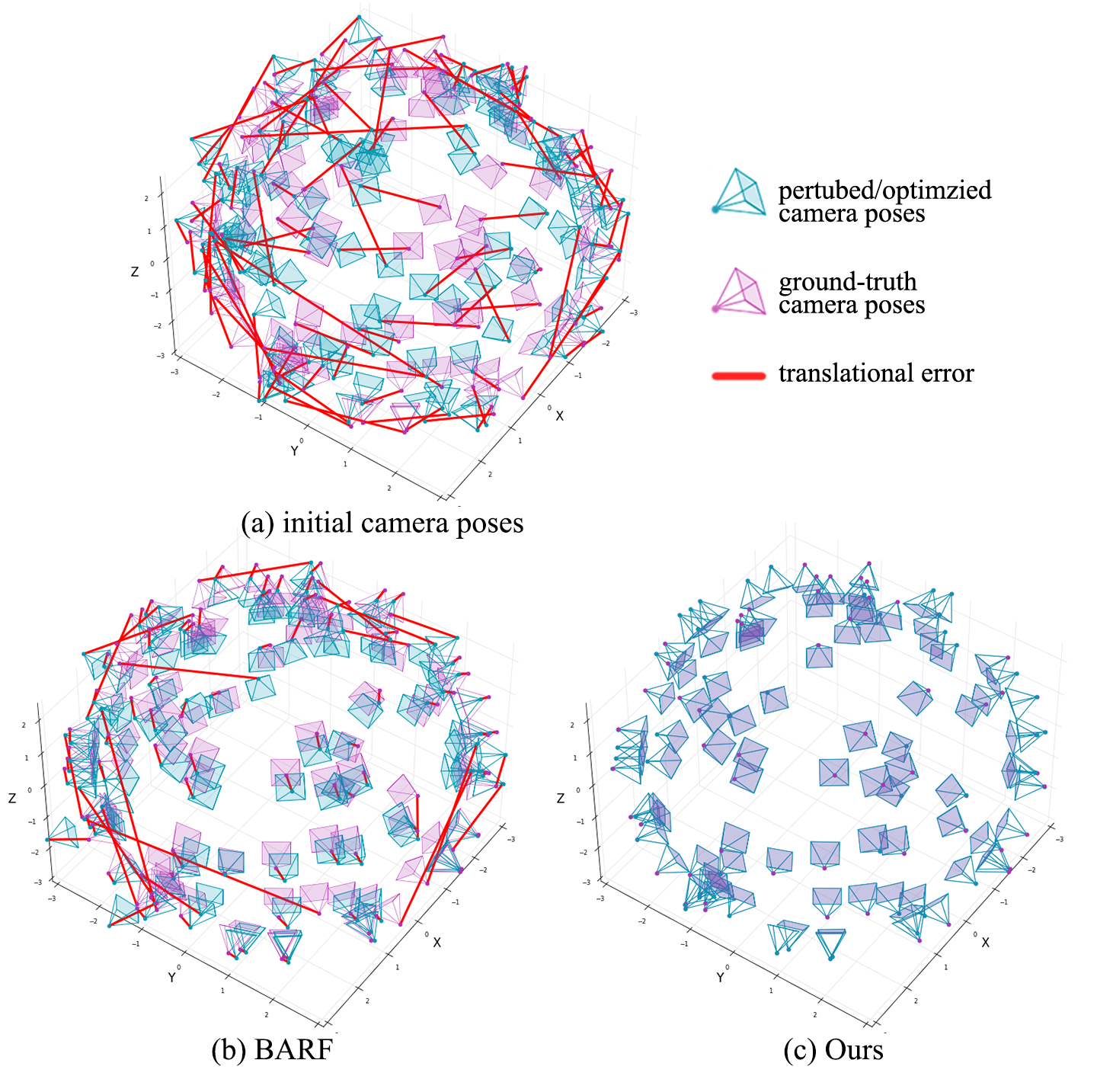}
	\vspace{-1em}
	\caption{The visualization of the pose optimization result. Each figure compares the ground-truth camera poses to the perturbed or optimized camera poses for the materials scene. The initial noise coefficient is set to 0.35. BARF encounters overfitting before completing the optimization, while CBARF successfully optimizes all camera poses. The camera poses are aligned by Procrustes Analysis.
} 
\label{pose}
\end{figure}

\begin{figure*}[!t]
	\centering
	\includegraphics[width=1.0\linewidth]{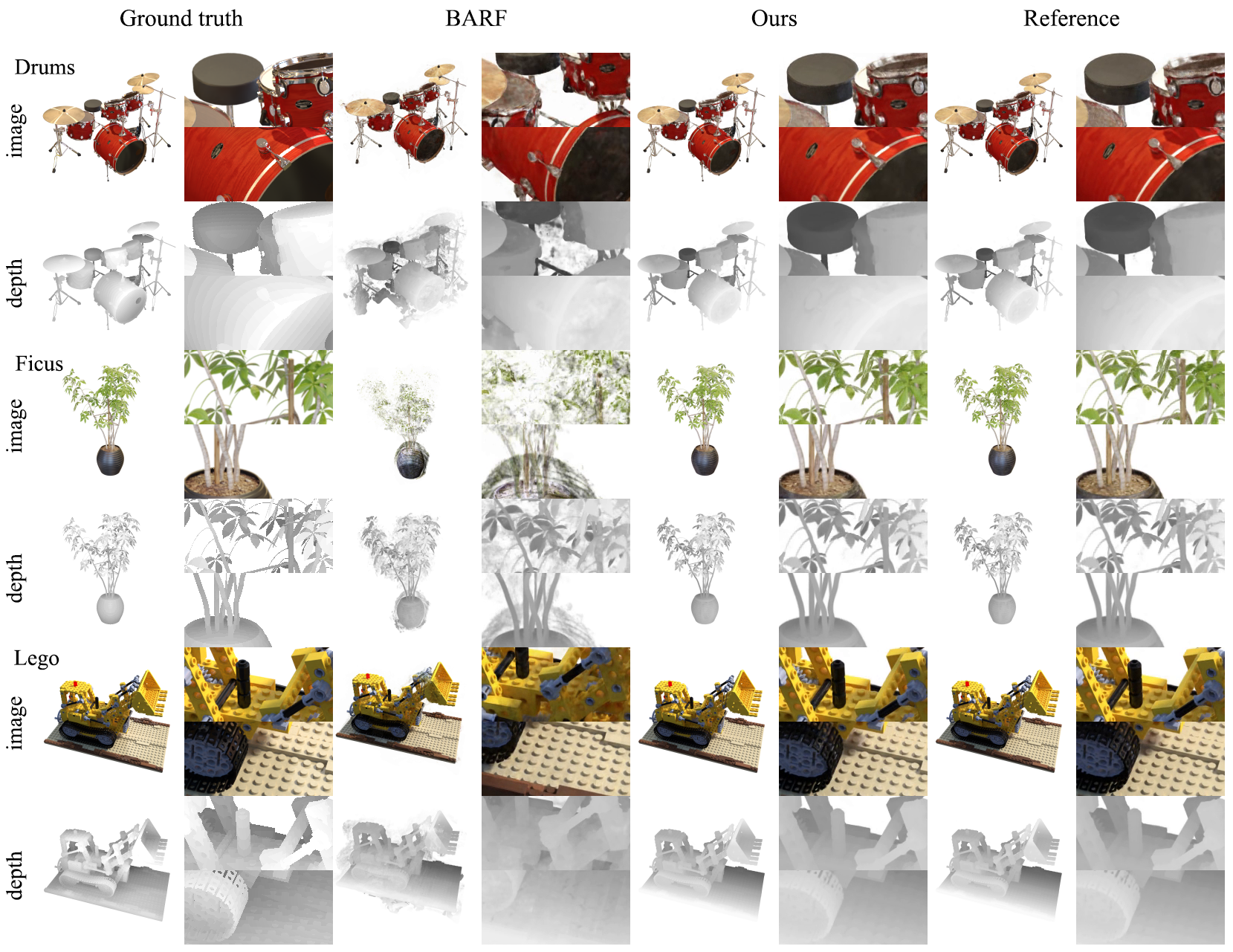}
	\vspace{-1em}
	\caption{Qualitative results of rendering on NeRF-synthetic scenes. For each scene, the top row displays the synthesized images, while the bottom row shows the estimated depth. To facilitate comparison, the reference images rendered at perfect camera poses are included on the rightmost column. CBARF achieves high-quality rendering results comparable to the reference rendered images, whereas BARF produces blurry and incorrect renderings due to unsuccessful camera pose optimization.
} 
\label{render1}
\end{figure*}
\begin{table*}[!t]
  \centering
  \caption{Quantitative results on synthetic object scenes. CBARF achieves superior performance from noisy camera poses compared to the baseline methods. Moreover, CBARF is able to maintain comparable view synthesis quality to the reference images rendered at the ground-truth camera poses. Translation errors are scaled by 100.}
    \begin{tabular}{c||cc|cc||cc|c|cc|c|cc|c}
    \hline
    \multirow{3}[4]{*}{Scene} & \multicolumn{4}{c||}{Camera pose optimization} & \multicolumn{9}{c}{View synthesis quality} \bigstrut[t]\\
         & \multicolumn{2}{c|}{Rotation (°)↓} & \multicolumn{2}{c||}{Translation↓} & \multicolumn{3}{c|}{PSNR↑} & \multicolumn{3}{c|}{SSIM↑} & \multicolumn{3}{c}{LPIPS↓} \bigstrut[b]\\
\cline{2-14}         & BARF & CBARF & BARF & CBARF & BARF & CBARF & ref. & BARF & CBARF & ref. & BARF & CBARF & ref. \bigstrut\\
    \hline
    Chair & 5.208 & \textbf{0.099} & 15.24 & \textbf{0.479} & 17.08 & \textbf{27.94} & 34.09 & 0.801 & \textbf{0.927} & 0.976 & 0.181 & \textbf{0.038} & 0.027 \bigstrut[t]\\
    Drums & 5.748 & \textbf{0.042} & 19.21 & \textbf{0.148} & 12.69 & \textbf{25.29} & 25.42 & 0.714 & \textbf{0.928} & 0.929 & 0.287 & \textbf{0.080} & 0.079 \\
    Ficus & 5.316 & \textbf{0.083} & 12.49 & \textbf{0.444} & 17.05 & \textbf{30.54} & 32.58 & 0.821 & \textbf{0.969} & 0.977 & 0.142 & \textbf{0.028} & 0.025 \\
    Hotdog & 4.931 & \textbf{0.248} & 14.30 & \textbf{1.305} & 15.97 & \textbf{23.44} & 36.76 & 0.827 & \textbf{0.891} & 0.980 & 0.223 & \textbf{0.078} & 0.033 \\
    Lego & 7.053 & \textbf{0.073} & 21.86 & \textbf{0.261} & 12.13 & \textbf{31.35} & 34.71 & 0.680 & \textbf{0.962} & 0.976 & 0.317 & \textbf{0.033} & 0.027 \\
    Materials & 11.85 & \textbf{0.047} & 28.68 & \textbf{0.179} & 11.09 & \textbf{28.90} & 29.58 & 0.669 & \textbf{0.947} & 0.950 & 0.311 & \textbf{0.061} & 0.059 \\
    Mic  & 6.568 & \textbf{0.063} & 17.26 & \textbf{0.252} & 13.45 & \textbf{30.56} & 33.11 & 0.827 & \textbf{0.976} & 0.982 & 0.172 & \textbf{0.020} & 0.018 \\
    Ship & 10.61 & \textbf{1.099} & 25.22 & \textbf{0.899} & 10.43 & \textbf{28.01} & 29.04 & 0.622 & \textbf{0.870} & 0.877 & 0.406 & \textbf{0.163} & 0.161 \bigstrut[b]\\
    \hline
    Mean & 7.161 & \textbf{0.219} & 19.28 & \textbf{0.496} & 13.74 & \textbf{28.25} & 31.91 & 0.745 & \textbf{0.934} & 0.956 & 0.255 & \textbf{0.062} & 0.054 \bigstrut\\
    \hline
    \end{tabular}%
  \label{table1}
\end{table*}%

\begin{figure*}[t]
	\centering
	\includegraphics[width=1.0\linewidth]{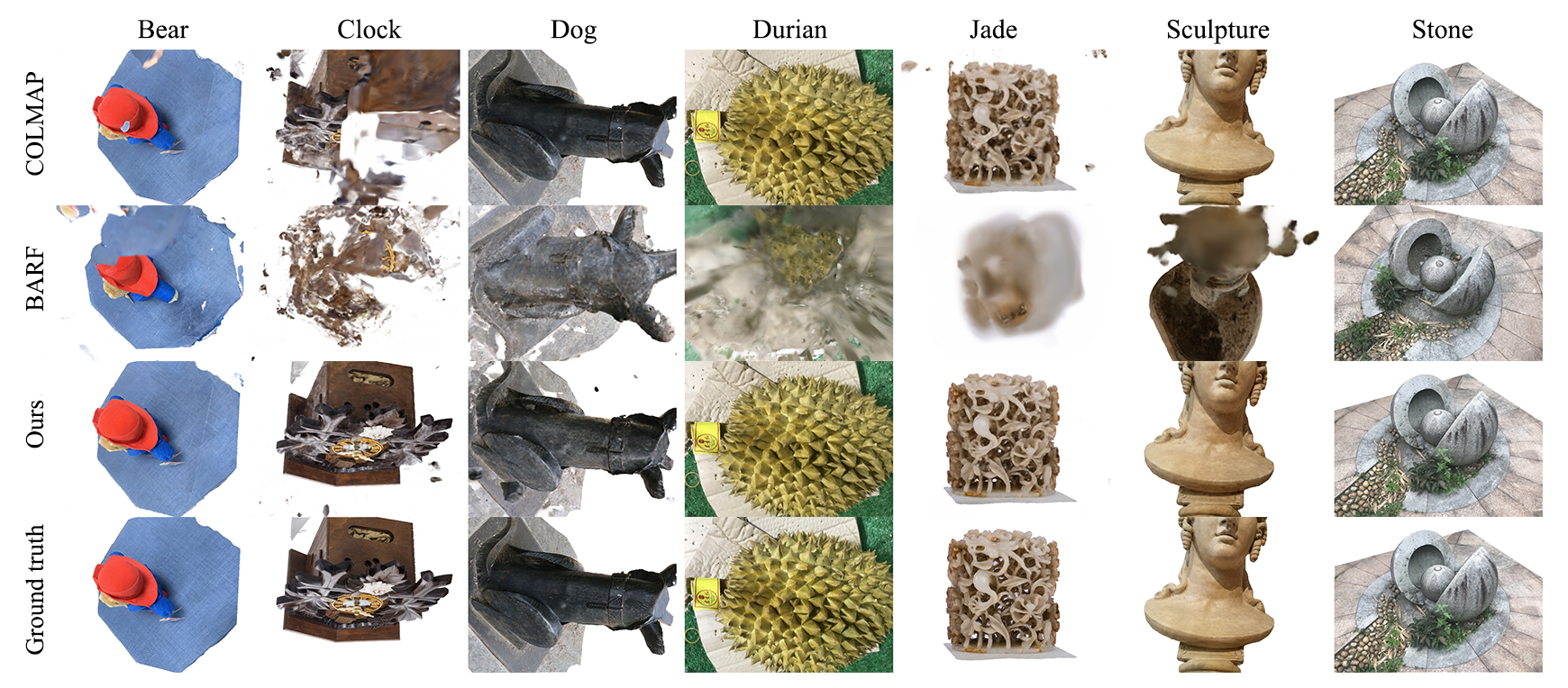}
	\vspace{-1em}
	\caption{The rendering results of inward-facing scenes using incomplete BlendedMVS datasets, with 10\% of the input images lacking camera pose information. We employ various models to estimate the missing camera poses and generate rendered images at these poses. A closer resemblance between the rendered image and the GT image indicates a more accurate estimation of the missing camera pose. BARF struggles to generate recognizable rendered images, while CBARF exhibits comparable view synthesis quality to the ground-truth images. In most scenes, CBARF also outperforms the reference model COLMAP.
} 
\label{render2}
\end{figure*}

\begin{table*}[htbp]

  \centering
  \caption{Quantitative evaluation of the rendered images obtained after optimizing the unknown camera poses. The rendering quality of images at the camera poses estimated by CBARF surpasses that of BARF and COLMAP. This indicates that CBARF successfully optimizes camera poses from incomplete datasets. Moreover, the camera poses estimated from CBARF are more accurate than those estimated by other methods. }
    \begin{tabular}{c||cc|c|cc|c|cc|c}
    \hline
    \multirow{3}[4]{*}{Scene} & \multicolumn{9}{c}{View synthesis quality} \bigstrut[t]\\
         & \multicolumn{3}{c|}{PSNR↑} & \multicolumn{3}{c|}{SSIM↑} & \multicolumn{3}{c}{LPIPS↓} \bigstrut[b]\\
\cline{2-10}         & BARF & CBARF & COLMAP & BARF & CBARF & COLMAP & BARF & CBARF & COLMAP \bigstrut\\
    \hline
    Bear & 10.64 & \textbf{23.85} & 21.68 & 0.525 & \textbf{0.718} & 0.698 & 0.583 & \textbf{0.318} & 0.339 \bigstrut[t]\\
    Clock & 8.75 & \textbf{16.68} & 9.74 & 0.443 & \textbf{0.672} & 0.539 & 0.593 & \textbf{0.415} & 0.542 \\
    Dog  & 10.11 & \textbf{19.43} & 19.25 & 0.377 & \textbf{0.680} & 0.669 & 0.597 & \textbf{0.341} & 0.351 \\
    Durian & 10.33 & 25.09 & \textbf{25.87} & 0.302 & 0.789 & \textbf{0.809} & 0.752 & 0.292 & \textbf{0.279} \\
    Jade & 11.57 & \textbf{22.31} & 17.54 & 0.681 & \textbf{0.812} & 0.684 & 0.424 & \textbf{0.234} & 0.375 \\
    Sculture & 11.06 & \textbf{29.11} & 26.86 & 0.691 & \textbf{0.934} & 0.915 & 0.400 & \textbf{0.110} & 0.126 \\
    Stone & 12.63 & \textbf{26.89} & 26.70 & 0.210 & \textbf{0.792} & 0.777 & 0.602 & \textbf{0.230} & 0.235 \bigstrut[b]\\
    \hline
    Mean & 11.38 & \textbf{24.23} & 21.75 & 0.503 & \textbf{0.793} & 0.749 & 0.522 & \textbf{0.250} & 0.294 \bigstrut\\
    \hline
    \end{tabular}%
  \label{table2}
\end{table*}%

\vspace{2mm}
\noindent \textbf{Evaluation Metrics.}
We evaluate the performance of our model in two main aspects: camera pose error for pose optimization and view synthesis quality for scene representation. 
For camera pose evaluation, we measure the rotation error and translation error separately to assess the accuracy of the optimized camera poses.
In terms of view synthesis evaluation, we employ several metrics including PSNR, SSIM and LPIPS \cite{zhang2018unreasonable} to provide quantitative measures of the similarity between the synthesized images and the reference images.

The reference camera poses are only used during the evaluation process to calculate the errors and are not used as supervision during the optimization process. As a result, the optimized camera poses may have a global offset from the ground-truth values.
The global offset can be thought of as the combination of overall rotation and translation. It does not affect the learning of scene representation, but it can introduce a misalignment between the coordinate system of the optimized camera poses and the ground-truth. This misalignment may lead to incorrect evaluations of pose optimization and scene reconstruction quality. Thus, we use Procrustes analysis \cite{gower2004procrustes} to align the optimized poses with the ground-truth before evaluation.
Procrustes analysis is a common technique used to align two sets of data points by minimizing the difference between them. By calculating the global offset through Procrustes analysis, we can align the optimized camera poses to the ground-truth and accurately calculate the rotation and translation errors of the optimized camera poses. Additionally, we can also align the camera poses in the test set for view synthesis evaluation.

\vspace{2mm}
\noindent \textbf{Results.}
We incorporate the camera poses with added noise as inputs to models and then perform optimization and reconstruction. We calculate both pose error and rendering quality. The visualization of the pose optimization result is shown in Fig. \ref{pose}. The results of rendering are visualized in Fig. \ref{render1} and the quantitative metrics are reported in Table. \ref{table1}.
As depicted in Table. \ref{table1}, CBARF consistently outperforms the baseline method BARF in terms of camera pose optimization. Additionally, our method demonstrates higher accuracy in estimating camera poses compared to the reference model COLMAP \cite{schonberger2016structure}.
The rendering results of our method, as illustrated in Figure \ref{render1}, exhibit comparable quality to the ground-truth images.

\subsection{Optimizing Incomplete Camera Poses}
\label{exp2}

\vspace{2mm}
\noindent \textbf{Experimental settings.}
In this work, we conduct evaluations on the BlendedMVS dataset \cite{yao2020blendedmvs}. The camera poses are estimated by COLMAP \cite{schonberger2016structure}.
To simulate the scenario where the camera pose estimation fails or is unavailable for certain images, we randomly drop 10\% of the camera poses and use the remaining images as the test set $F$. The other images with camera poses are grouped as $T$. We then assign an initial camera pose $\mathbf{P}_{ini}$ to the images in $F$.
During the training phase, we use both BARF and CBARF to learn scene representation from $T$ and jointly optimize the camera poses of both groups, resulting in $T'$ and $F'$. 
Since there are no ground-truth camera poses for evaluation, we use optimized poses in $F'$ to assess rendering quality to estimate the camera pose error.
In the testing phase, we compare the rendering results generated using the camera poses estimated by different models. This allows for a meaningful comparison of the rendering quality between different approaches.

\vspace{2mm}
\noindent \textbf{Results.}
Due to the lack of ground-truth camera poses, it is not possible to calculate the error of camera poses. We can only calculate the rendered image quality to evaluate the performance of different models.
The results on the BlendedMVS dataset are visualized in Fig. \ref{render2}, and the quantitative rendering quality is reported in Table. \ref{table2}. We use the quality of rendered images as an indirect measure to evaluate the pose optimization capability of different models, as higher rendering quality highly probably indicates more accurate camera pose estimation.



\section{Conclusions}

In this paper, we propose CBARF (Cascaded Bundle-Adjusting Neural Radiance Fields), a novel 3D reconstruction model aiming to effectively optimize imperfect camera poses. 
We demonstrate the significance of camera pose initialization for the performance of bundle-adjustment (BA). Consequently, we introduce the cascaded BA to progressively refine the camera poses.
Then our proposed neighbor-replacement strategy effectively rectifies erroneous poses that cannot be automatically optimized in the BA process. We also design a novel criterion to identify such poorly estimated poses without relying on ground-truth.
Our experiments demonstrate the superiority of our CBARF model in both camera pose optimization and novel view synthesis.

Despite the success of our methods, there are still areas for improvement. CBARF has similar limitations to the reference model BARF \cite{lin2021barf}, including rigidity assumption and dependence on initial pose information. Future research can explore extensions of the CBARF model for more complex scenes and reduced reliance on initial pose information. We believe CBARF opens up new possibilities for 3D reconstruction framework with unknown camera poses.





\bibliographystyle{IEEEtran}
\bibliography{egbib}

\vspace{11pt}


\end{document}